\documentclass{Interspeech2024}

\usepackage{multirow}

\interspeechcameraready

\title{HuBERT-EE: Early Exiting HuBERT for Efficient Speech Recognition}

\name[affiliation={1}]{Ji Won}{Yoon}
\name[affiliation={2}]{Beom Jun}{Woo}
\name[affiliation={2}]{Nam Soo}{Kim}

\address{
  $^1$Department of AI, Chung-Ang University, Seoul, South Korea\\
  $^2$Department of ECE and INMC, Seoul National University, Seoul, South Korea}
\email{jiwonyoon@cau.ac.kr, bjwoo@hi.snu.ac.kr, nkim@snu.ac.kr}


\begin{document}

\maketitle

\begin{abstract}
    Pre-training with self-supervised models, such as Hidden-unit BERT (HuBERT) and wav2vec 2.0, has brought significant improvements in automatic speech recognition (ASR). However, these models usually require an expensive computational cost to achieve outstanding performance, slowing down the inference speed. To improve the model efficiency, we introduce an early exit scheme for ASR, namely HuBERT-EE, that allows the model to stop the inference dynamically. In HuBERT-EE, multiple early exit branches are added at the intermediate layers. When the intermediate prediction of the early exit branch is confident, the model stops the inference, and the corresponding result can be returned early. We investigate the proper early exiting criterion and fine-tuning strategy to effectively perform early exiting. Experimental results on the LibriSpeech show that HuBERT-EE can accelerate the inference of the HuBERT while simultaneously balancing the trade-off between the performance and the latency.
\end{abstract}

\section{Introduction}
Recently, self-supervised speech representation learning (speech SSL) \cite{w2v,hubert,wavlm, ils-ssl, data2vec} has achieved considerable improvements in automatic speech recognition (ASR) literature.
Unlike fully-supervised learning approaches, which rely on manually annotated labels, speech SSL models can learn a meaningful speech representation by leveraging unlabeled speech data.

Among the various speech SSL models, Hidden-unit BERT (HuBERT) \cite{hubert} is one of the most prominent models for speech recognition.
On the LibriSpeech \cite{librispeech}, fine-tuned HuBERT using connectionist temporal classification (CTC) \cite{ctc} achieves the state-of-the-art word error rate (WER) results.
However, such a model tends to have a large model size and high computational complexity to achieve promising performance.
For example, the base version of the HuBERT has about 95 million parameters.
Also, a large version utilizes twice as many Transformer layers \cite{transformer} as in the base version, with almost 317 million parameters.
These large-scale pre-trained models usually suffer from slow inference speed, which may hinder their usage in real-world applications where fast inference is desirable.

Typical approaches to improve model efficiency include knowledge distillation (KD) \cite{kd, bucila-et-al:scheme}, pruning \cite{pruning1, pruning2}, and model quantization \cite{quantization}. 
While those methods reduce the processing complexity, they still require samples to pass through the entire model.
In contrast, early exiting is a technique to adaptively accelerate the inference speed by returning the result at an intermediate layer.
Since multiple classifiers are attached to some intermediate layers and jointly trained with the original backbone model, each classifier yields the prediction and confidence score during the inference.
When the intermediate prediction is confident enough, the corresponding result can be exited early.
However, existing early exit methods \cite{ee1,ee2,ee3,ee4,ee5} are mainly designed for natural language processing (NLP) classification tasks.
Only a few studies have been investigated in the speech domain, including speech enhancement \cite{ee-se}, speech separation \cite{ee-sp}, and limited-vocabulary commands recognition \cite{ee-cr}.
Since the ASR model does not use the commonly-used classifier for classification, it is challenging to directly apply the previous approaches to the ASR model.

In this paper, we introduce a simple yet effective early exit method for ASR, namely HuBERT Early Exiting (HuBERT-EE), that enables the HuBERT model to stop the inference dynamically.
\emph{To the best of our knowledge, this is the first attempt to apply the early exit framework to the speech SSL model.}
Specifically, the proposed HuBERT-EE accelerates the inference procedure by adding multiple early exit branches at the intermediate layers of the HuBERT.
When the early exit branch is confident in its prediction, the model stops the inference and outputs the intermediate prediction as the final result.
Different from intermediate CTC-based approaches \cite{inter-ctc, inter-ctc2, inter-ctc-pruning, inter-KD}, the HuBERT-EE aims to dynamically use the intermediate prediction as the model's final output with minimal WER degradation.
Instead of simply applying the intermediate-CTC framework, we newly construct the self-attention-based early exit branch to perform the early exiting effectively.
In addition, we explore the proper early exiting criterion and fine-tuning strategy to perform the early exiting effectively. Also, we newly design the self-attention-based early exit branch.

From the experimental results on the LibriSpeech dataset, it is verified that the HuBERT-EE can be successfully applied to the ASR task.
Compared to the other compression methods, HuBERT-EE enables the model to stop the inference dynamically while achieving a better speed-performance trade-off.
This implies that the proposed method can be applied to a real-world scenario in which users have the flexibility to adjust the inference speed according to their demands.

\begin{figure}[t]
    \centering
        \includegraphics[height=6.2cm]{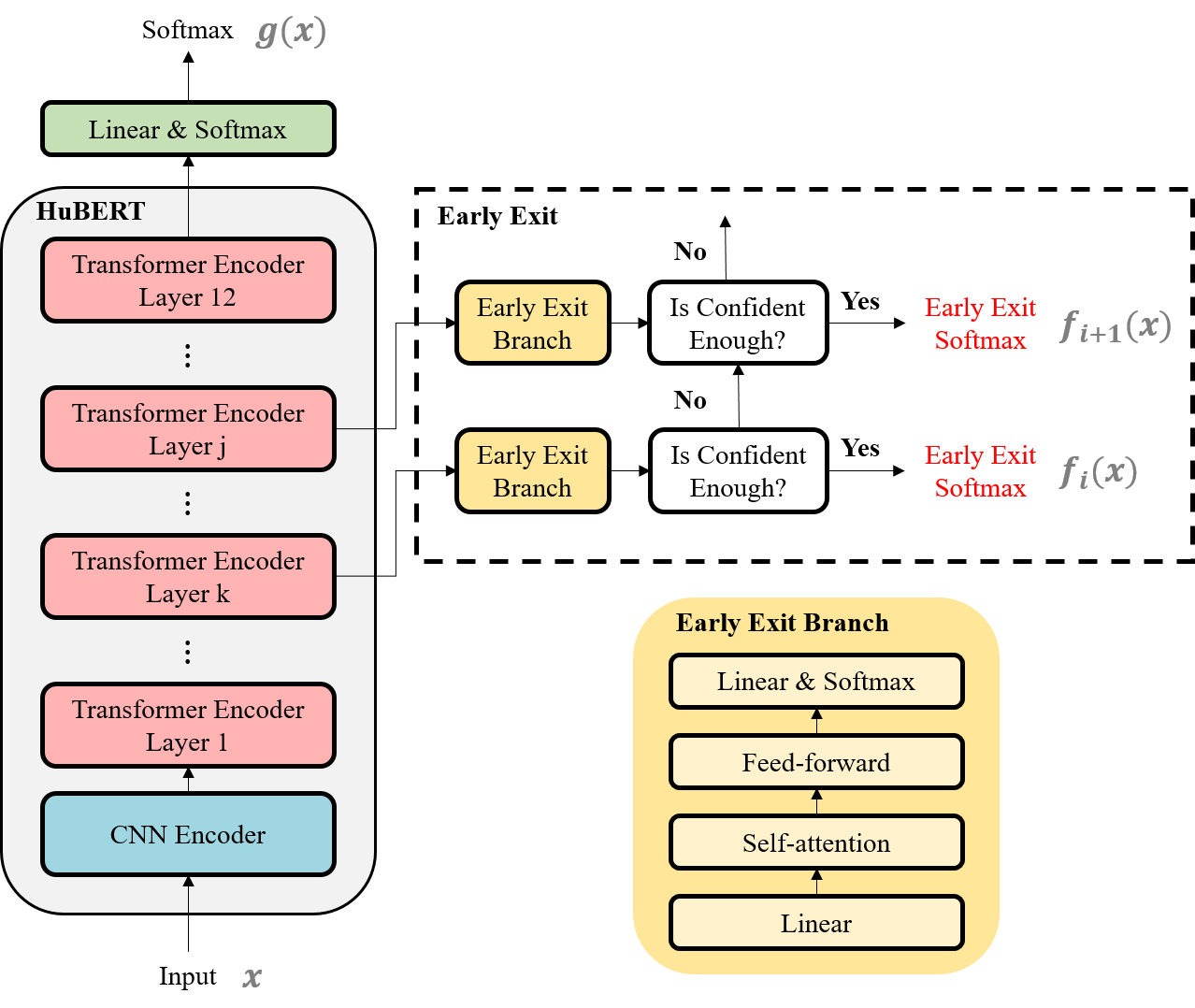}
    \caption{Overview of the HuBERT-EE. 
    In contrast to the original HuBERT model, our proposed approach can stop the inference dynamically. 
    If an early exit branch is sufficiently confident in its prediction, the corresponding result can be exited early. 
    }
  \label{proposed}
\end{figure}

\section{HuBERT-EE}

Conventional early exit methods have mainly focused on NLP classification tasks. Considering that the ASR model does not employ the typical classifier used in classification tasks, it is essential to develop a new early exiting framework specifically tailored for ASR. In this section, we introduce a pioneering early exit method designed for HuBERT, namely HuBERT-EE. Although our experiments utilize HuBERT as the backbone model, the proposed framework can be extended to other SSL models as well.

\subsection{Model architecture}

As shown in Figure \ref{proposed}, HuBERT-EE mainly consists of a backbone network and multiple early exit branches.
The backbone is built upon a 12-layers Transformer encoder with an additional linear projection layer.
The early exit branches are located at some intermediate layers of the backbone network to enable early predictions.

Following the previous studies \cite{ee1,ee2}, we apply the base version of the HuBERT as for the backbone, which contains 12 Transformer encoder layers.
For the convenience of notation, we let HuBERT-base denotes the base version of the HuBERT.
The backbone is composed of a convolutional neural network (CNN) encoder, 12 Transformer encoder layers, and the linear layer with softmax function. 
The structure of the CNN encoder and the Transformer encoder conform with those of the original HuBERT.
In order to fine-tune the model for the speech recognition task, the linear projection layer is added to the final layer of the HuBERT-base model.

In the original HuBERT, the way to fine-tune the model is to attach one linear projection layer to the final Transformer encoder layer.
Then, the linear layer outputs the prediction for the ASR task.
In the proposed method, we add early exit branches to the intermediate layers of the HuBERT, enabling more efficient CTC predictions, like intermediate-CTC \cite{inter-ctc}.
However, we experimentally found that simply applying intermediate-CTC with a linear layer did not perform well in the early exit framework.
Instead, as shown in Figure \ref{proposed}, we newly construct the early exit branch with the self-attention layer at its core, motivated by the Transformer structure.
It is designed carefully to balance the trade-off between performance and efficiency.

\subsection{Pre-training}
Before the fine-tuning stage, we start with pre-training the backbone model on unlabelled data with a self-supervised learning objective.
This stage is identical to the vanilla HuBERT pre-training.
Note that the linear layer, located at the final Transformer layer, and all early exit branches stay unaffected during the pre-training.
In our experiments, we used the pre-trained HuBERT checkpoint, which is provided by  the Fairseq \cite{fairseq} toolkit.

\subsection{Fine-tuning}
\label{fine-tuning-sec}
In this subsection, we discuss how to fine-tune the proposed HuBERT-EE.
Firstly, the CTC loss function for the last linear projection layer, which is located on top of the Transformer, can be formulated as 
\begin{align}
  \mathcal{L}_{FT1} = CTC_{loss}(y,g(x))
\end{align}%
where $x$, $y$, $g$, $CTC_{loss}$ denote the input sequence, the corresponding label, the output of the linear projection layer, and the CTC loss, respectively.
This training is identical to the HuBERT fine-tuning in the original paper \cite{hubert}.

For fine-tuning the early exit branches on the ASR task, the CTC loss function of the $i^{th}$ early exit branch is as follows:
\begin{align}
  \mathcal{L}_{i} = CTC_{loss}(y,f_{i}(x))
\end{align}%
where $f_{i}$ denotes the output of the $i^{th}$ early exit branch.
When there are $N$ branches in the HuBERT-EE, the loss for fine-tuning all early exit branches can be calculated as
\begin{align}
  \mathcal{L}_{FT2} = \sum_{i=1}^{N}\mathcal{L}_{i}.
\end{align}%
Due to the performance degradation, we consider uniform weights for training the early exit branches instead of the weighted average \cite{weightedaverage}.

Based on the two losses $\mathcal{L}_{FT1}$ and $\mathcal{L}_{FT2}$, we investigate the effective fine-tuning approach to train the HuBERT-EE.
In the previous related studies, there are mainly two fine-tuning strategies for training the early-exit model: (1) joint training that jointly fine-tunes the final linear layer and all early-exit branches and (2) two-stage training that fine-tunes the two components separately. In Section \ref{proper_sec}, we compare these two fine-tuning strategies and look for the proper one to train our framework.

The straightforward fine-tuning approach is to jointly train the last linear layer and all the early exit branches \cite{ee3, ee4} by minimizing the sum of the two loss functions $\mathcal{L}_{FT1} + \lambda \mathcal{L}_{FT2}$. In our experiments, we experimentally set $\lambda$ to 1.

When it is required to maintain the best performance of the final linear layer, the two-stage training is the desired fine-tuning approach \cite{ee1}.
In this training scheme, we first fine-tune the whole model weights with the loss function $\mathcal{L}_{FT1}$, except for the early exit branches.
Then, we freeze all parameters fine-tuned in the previous stage and only update the early exit branches with CTC loss $\mathcal{L}_{FT2}$.
Note that the reason for freezing parameters of the backbone and the final linear layer is to keep the high performance of the original HuBERT-base.

\subsection{Early exit inference}
After fine-tuning HuBERT-EE for ASR, the model is capable of making early exit decisions during the inference procedure.
Each early exit branch, added at the intermediate Transformer layer, outputs the prediction and confidence score.
If the intermediate prediction is confident enough, the forward inference is terminated, and the result is returned early.
In this paper, we quantify the early exit branch’s confidence in its prediction in two ways: entropy and maximum probability.
In Section \ref{exp_ee_criterion}, we compare these two criteria and determine the optimal one.

\subsubsection{Entropy}
Since entropy is a well-known measure of uncertainty, we use the entropy-derived confidence measure as the early exit criterion.
The entropy of the $i^{th}$ early exit branch’s output $f_{i}(x)$ can be computed as
\begin{align}
    \text{Entropy} = -{1 \over {T \times C}}\sum_{T}\sum_{C}f_{i}(x)\times \log{f_{i}(x)}.
    \label{entropy}
\end{align}%
The prediction with lower entropy might be more confident to exit.
If the entropy of $f_{i}(x)$ is lower than the preset threshold $S$, HuBERT-EE stops the inference, returning the result early.

\subsubsection{Confidence}
The maximum probability is another straightforward measure of certainty.
Since we use the CTC framework, the softmax prediction of the $i^{th}$ early exit branch can be expressed as $f_{i}(x) \in R^{T \times C}$, where $T$ is the total number of frames and $C$ is the number of label classes. 
Considering the maximum probability as the confidence measure, the average confidence score of $f_{i}(x)$ is given as
\begin{align}
\begin{small}
  \text{Confidence} = {1 \over T} \sum_{T}\max_{c}{f_{i}(x)}^{(c)}
  \label{confidence}
  \end{small}
\end{align}
where $\max_{c}{f_{i}(x)}^{(c)} \in R^{T\times1}$ represents the maximum probability for each frame.
When the confidence of the intermediate output $f_{i}(x)$ is larger than the predefined threshold, the corresponding prediction can be exited early.

\section{Experiments}

\subsection{Experimental setup}

We used the LibriSpeech \cite{librispeech} (about 1000 hours) for pre-training and supervised fine-tuning.
As the training dataset, “train-clean-100”, “train-clean-360”, and “train-other500” were used. 
For validation, we used ``dev-other".
We applied ``test-clean” and ``test-other” for evaluation.

We applied HuBERT-EE to the HuBERT-base model, containing 12 Transformer encoder layers.
For implementation, the Fairseq \cite{fairseq} toolkit was mainly utilized to build the models.
In the case of the early exit branch, we added early exit branches to three layers: $5^{th}$, $8^{th}$, and $11^{th}$ layers of the HuBERT-base.
Each early exit branch module had the self-attention dimension $D_{EE}$ of 512 with four heads.
The additional early exit branches corresponds to about 22 M parameters, resulting in a total of 116 M parameters for the HuBERT-EE.
Instead of directly pre-training the HuBERT backbone model, we used the pre-trained checkpoint, provided by the Fairseq toolkit.
When fine-tuning the HuBERT-EE, we followed the fine-tuning scheme of the original paper \cite{hubert}, and the training was performed on four NVIDIA Quadro RTX 8000 GPUs.

We compared the HuBERT-EE with other compression techniques, including DistilHuBERT \cite{distillhubert} and LayerDrop \cite{layerdrop}.
All the models were pre-trained and fine-tuned using 960 hours of LibriSpeech.
To fairly compare the results, a single linear layer was employed as the ASR module, placed on top of the SSL model.
Both the pre-trained SSL model and the ASR module were fine-tuned together during the training.
We found that the original DistilHuBERT model, which consisted of two Transformer encoder layers, did not perform well when using a linear layer as the ASR module.
To address this, we experimented with 8 Transformer encoder layers of DistilHuBERT, referred to as DistilHuBERT-8L. 
Regarding LayerDrop, we applied it to the HuBERT-base model during the fine-tuning procedure and set the LayerDrop rate $p$ to 0.1 and 0.3.

We measured two performance metrics: word error rate (WER) and real time factor (RTF). WER is a widely used metric to evaluate the accuracy of ASR task, and RTF measures a decoding speed with the ratio between the ASR processing time and the utterance duration.

For the inference, we applied greedy decoding without a language model.
Since some model configurations did not support CPU-only inference, we evaluated GPU-based inference for each model.
The ASR models with a large model size typically use GPU resources for inference, so it is reasonable to utilize the GPU for decoding models.
RTF was measured on a single NVIDIA Quadro RTX 8000 GPU with single batch size, and we averaged RTF results over three runs.

\subsection{Exploring suitable early exiting criterion}
\label{exp_ee_criterion}

\begin{table}[t]
\centering
{\fontsize{7.5}{8.5}\selectfont
\begin{tabular}{cccc}
\hline
\multicolumn{2}{c}{\multirow{2}{*}{}}                                                                  & \multicolumn{2}{c}{test-clean} \\ \cline{3-4} 
\multicolumn{2}{c}{}                                                                                   & WER       & RTF ($\times 10^{-3}$)       \\ \hline
\multicolumn{2}{c}{HuBERT-base (backbone)}                                                               & 3.88 \%     & 3.529    \\ \hline
\multirow{6}{*}{\begin{tabular}[c]{@{}c@{}}HuBERT-EE\\ (Ours)\end{tabular}} & Entropy Thres.=0.0040  & 8.05 \%        & 2.879\\
                                                                            & Entropy Thres.=0.0035 & 6.50 \%        & 2.999      \\
                                                                            & Entropy Thres.=0.0025 & 4.17 \%        & 3.312 \\
                                                                            \cline{2-4} 
                                                                            & Confidence Thres.=0.950 & 8.25 \%        & 2.887 \\
                                                                            & Confidence Thres.=0.955 & 6.82 \%        & 3.043\\
                                                                            & Confidence Thres.=0.960 & 5.62 \%        & 3.308 \\ \hline
      \end{tabular}}
\caption{WER (\%) on test-clean dataset using different predefined thresholds.}
\vspace{-4mm}
\label{criterion_table}
\end{table}
To determine the proper early exit criterion, we examined the predefined threshold values for both entropy (in Eq. (\ref{entropy})) and confidence (in Eq. (\ref{confidence})). 
As shown in Table \ref{criterion_table}, we observed that both confidence and entropy-derived criterions performed well on ASR.
However, the entropy criterion was more supportive in making early exit decisions, achieving better WER performance with lower RTF values.
Specifically, on the test-clean dataset, HuBERT-EE with the entropy threshold 0.0035 achieved a WER of 6.50 \%. In contrast, HuBERT-EE with the confidence threshold 0.955 resulted in a slightly worse WER of 6.82 \%, while exhibiting slower inference speed. 
This suggests that utilizing the entropy criterion in HuBERT-EE leads to better trade-offs between WER and inference speed compared to the confidence one.
Therefore, we applied the entropy-derived criterion in Eq. (\ref{entropy}) as the baseline metric to decide the exiting.

\begin{figure}[t]
\centering
\vspace{-3mm}
	\includegraphics[height=2.8cm]{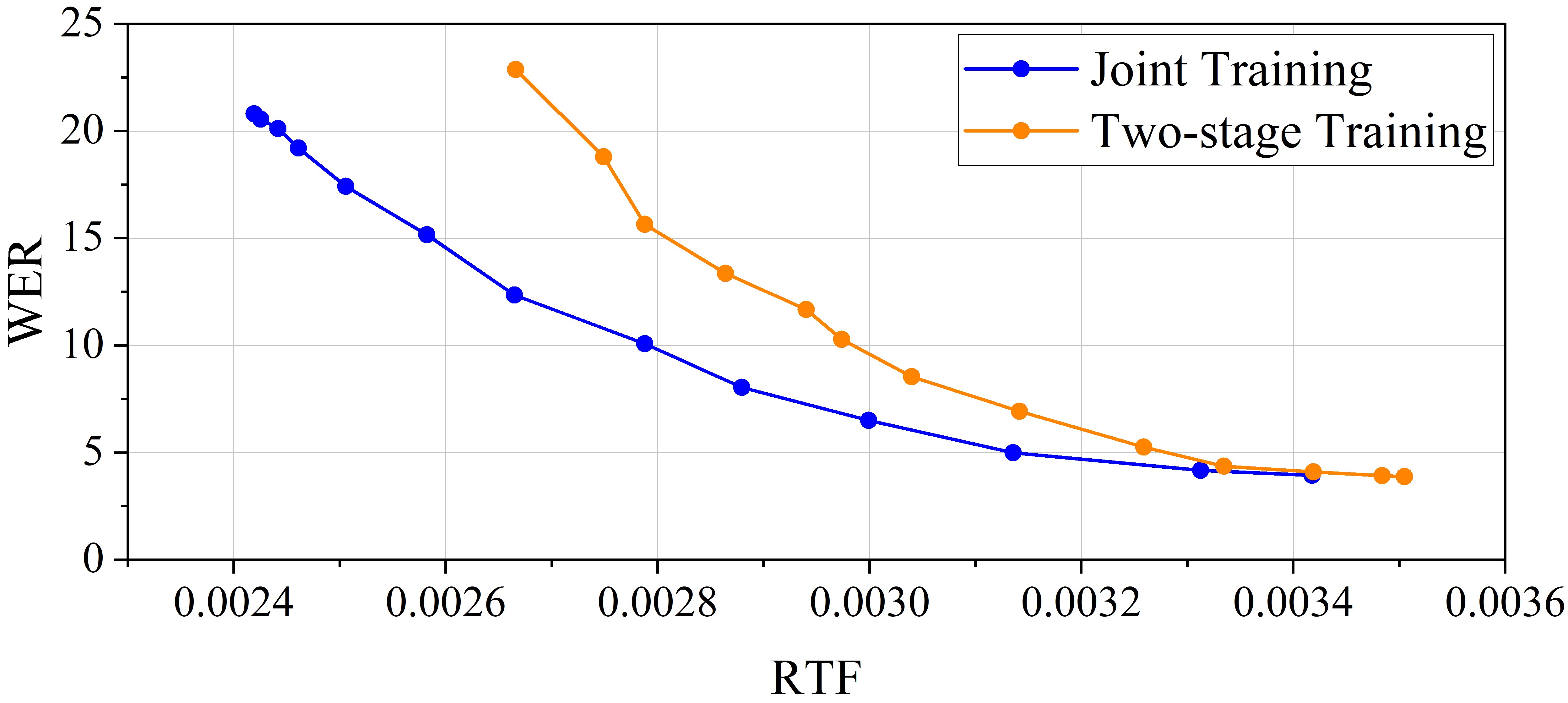}
 \vspace{-3mm}
	\caption{
    Quality–efficiency trade-offs on test-clean dataset using different fine-tuning strategies for HuBERT-EE. We set entropy thresholds $S$ from 0.008 to 0.002.}
 \vspace{-3mm}
	\label{joint_vs_two_stage}
\end{figure}

\begin{table}[t]
\centering
{\fontsize{7.5}{8.5}\selectfont
\begin{tabular}{ccc}
\hline
Exit Layer & Joint Training & Two-stage Training \\ \hline
5          & 21.11 \%         & 37.36 \%              \\ 
8          & 8.60 \%            & 11.99 \%              \\ 
11         & 4.04 \%           & 4.21 \%               \\ 
12         & 3.90 \%           & 3.88 \%               \\ \hline
\end{tabular}
}
\caption{Each exit layer's WER (\%) on test-clean dataset using different fine-tuning strategies.}
\vspace{-3mm}
\label{joint_vs_freeze_table}
\end{table}

\subsection{Proper fine-tuning strategy for HuBERT-EE}
\label{proper_sec}
In Section \ref{fine-tuning-sec}, we discussed two fine-tuning strategies for HuBERT-EE: (1) joint training and (2) two-stage training.
In Figure \ref{joint_vs_two_stage}, we visualized the trade-off while setting different entropy thresholds from 0.008 to 0.002.
Entropy is adopted as the early exit criterion, as it performed better than confidence in the previous experiment.
We measured both RTF and WER performance on the test-clean. The trade-off curves demonstrate that the HuBERT-EE with joint training showed a better trade-off compared to the two-stage training approach.
As the RTF value decreases, the difference between the two methods became more apparent.
This is because the two-stage training was considerably weaker in earlier layers, as shown in Table \ref{joint_vs_freeze_table}.
From the results, it is confirmed that the joint training was preferable as the fine-tuning strategy of the HuBERT-EE.

\begin{figure}[t]
\centering
        \vspace{-5mm}
	\includegraphics[height=3.3cm]{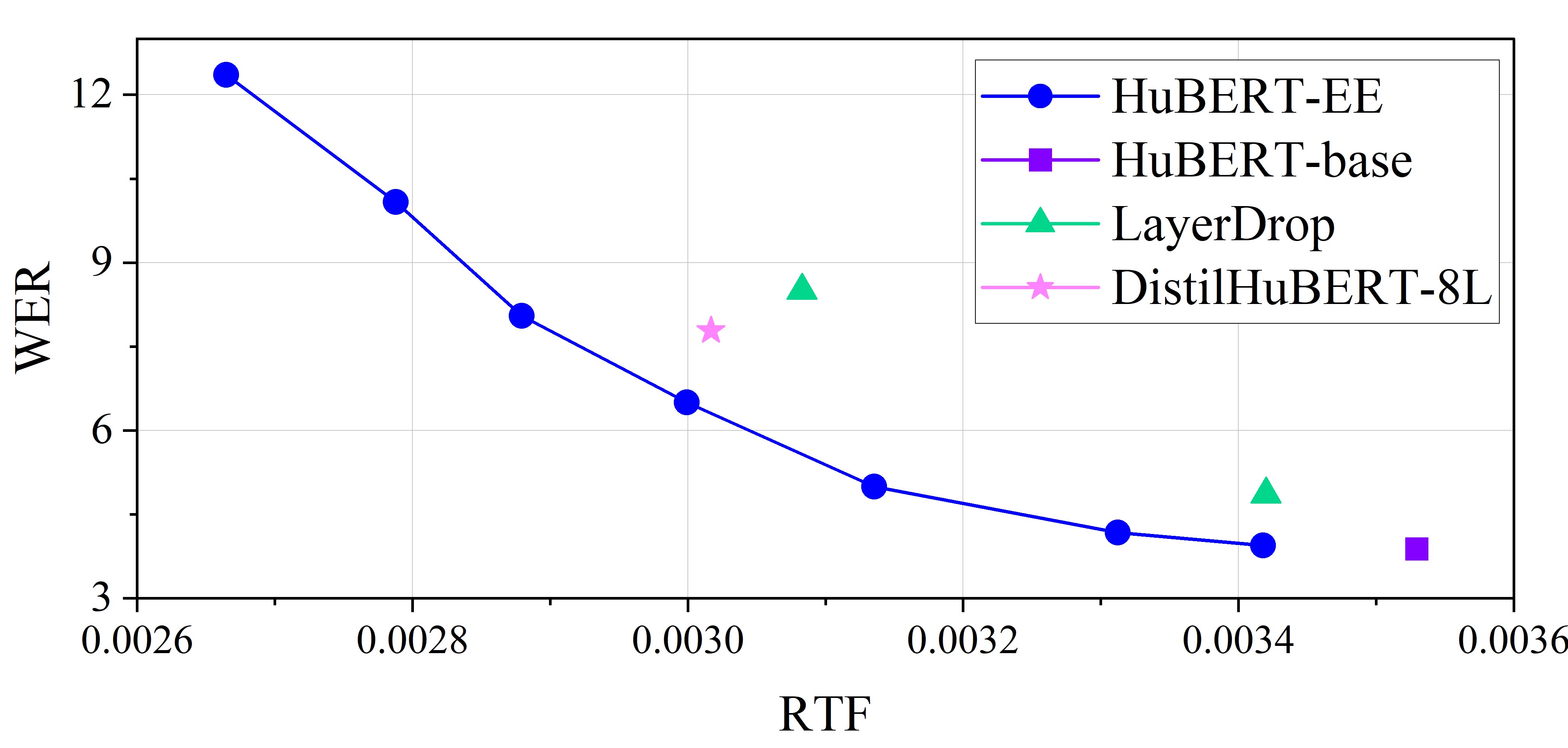}
        \vspace{-3mm}
	\caption{
    Performance comparison on test-clean. All results were evaluated based on greedy decoding. We set different thresholds $S$ from 0.005 to 0.002 for HuBERT-EE. The proposed model was fine-tuned with joint training.}
     \vspace{-1mm}
	\label{trade_other}
\end{figure}

\begin{table}[t]
\centering
{\fontsize{7.5}{8.5}\selectfont
\begin{tabular}{ccccc}
\hline
\multicolumn{2}{c}{\multirow{2}{*}{}}                                                                  & \multicolumn{3}{c}{test-other}                     \\ \cline{3-5} 
\multicolumn{2}{c}{}                                                                                   & \multicolumn{1}{c}{WER}  & \multicolumn{1}{c}{RTF ($\times 10^{-3}$)} & Speed\\ \hline
\multicolumn{2}{c}{HuBERT-base (backbone)}                                                               & \multicolumn{1}{c}{9.09 \%  } & 3.629  & 42.12 Hz             \\ \hline
\multicolumn{2}{c}{DistilHuBERT-8L}                                                                      & \multicolumn{1}{c}{19.21 \% }     & 3.023 & 50.55 Hz

                \\ \hline
\multirow{2}{*}{LayerDrop}                                                  & $p$=0.1                    & 10.93 \%                     & 3.493 & 43.75 Hz               \\
                                                                            & $p$=0.3                    & 16.92 \%                      & 3.085 
                                                                            & 49.54 Hz \\ \hline
\multirow{3}{*}{\begin{tabular}[c]{@{}c@{}}HuBERT-EE\\ (Ours)\end{tabular}} & S=0.0055  & 18.20 \%
                     & 2.955 & 51.72 Hz                \\
                                                                            & S=0.005 & 16.13 \%                      & 3.043    & 50.23 Hz           \\
                                        
                                                                        & S=0.003  & 10.04 \%                     & 3.439  
                                                                        & 44.44 Hz \\ \hline
\end{tabular}
}
\caption{Performance comparison on test-other. Speed of $k$ Hz means that the model can process $k$ samples per second.}
\vspace{-3mm}
\label{main_result}
\end{table}

\subsection{Performance comparison with conventional methods}
We compared the performance of the proposed approach with the conventional compression methods for HuBERT, including DistilHuBERT \cite{distillhubert} and LayerDrop \cite{layerdrop}.
DistilHuBERT is the recent distillation method to reduce the size of HuBERT, and LayerDrop is an effective structured pruning technique for Transformer network.
We used the entropy-based metric as the early exit criterion and fine-tuned HuBERT-EE with joint training due to their promising results in previous experiments. 
Figure \ref{trade_other} shows quality–efficiency trade-offs on test-clean.
From the results, it is verified that HuBERT-EE indeed achieved a better speed-performance trade-off compared to the others.
In addition, the proposed framework enabled HuBERT to adjust the inference speed without requiring model retraining. This flexibility is particularly advantageous in resource-constrained scenarios. By fine-tuning specific early exit branches, HuBERT-EE could provide greater control over the inference speed.
\emph{It's important to note that small RTF gains were a result of our GPU-based evaluation since some baseline model configurations did not support CPU-only inference.}
The technique's significance goes beyond RTF gains. HuBERT-EE outperformed DistilHuBERT and LayerDrop, allowing the model to stop the inference dynamically.
As shown in Table \ref{main_result}, HuBERT-EE still performed better on the test-other dataset.
Overall, the experimental results suggest that HuBERT-EE could be a promising solution for efficient ASR inference.
It striked a favorable balance between performance and efficiency, making it an attractive choice for practical ASR applications.

\subsection{Number of exiting samples}
\label{ex_layer}
\begin{figure}[t]
\centering
    \vspace{-3mm}
	\includegraphics[height=2.5cm]{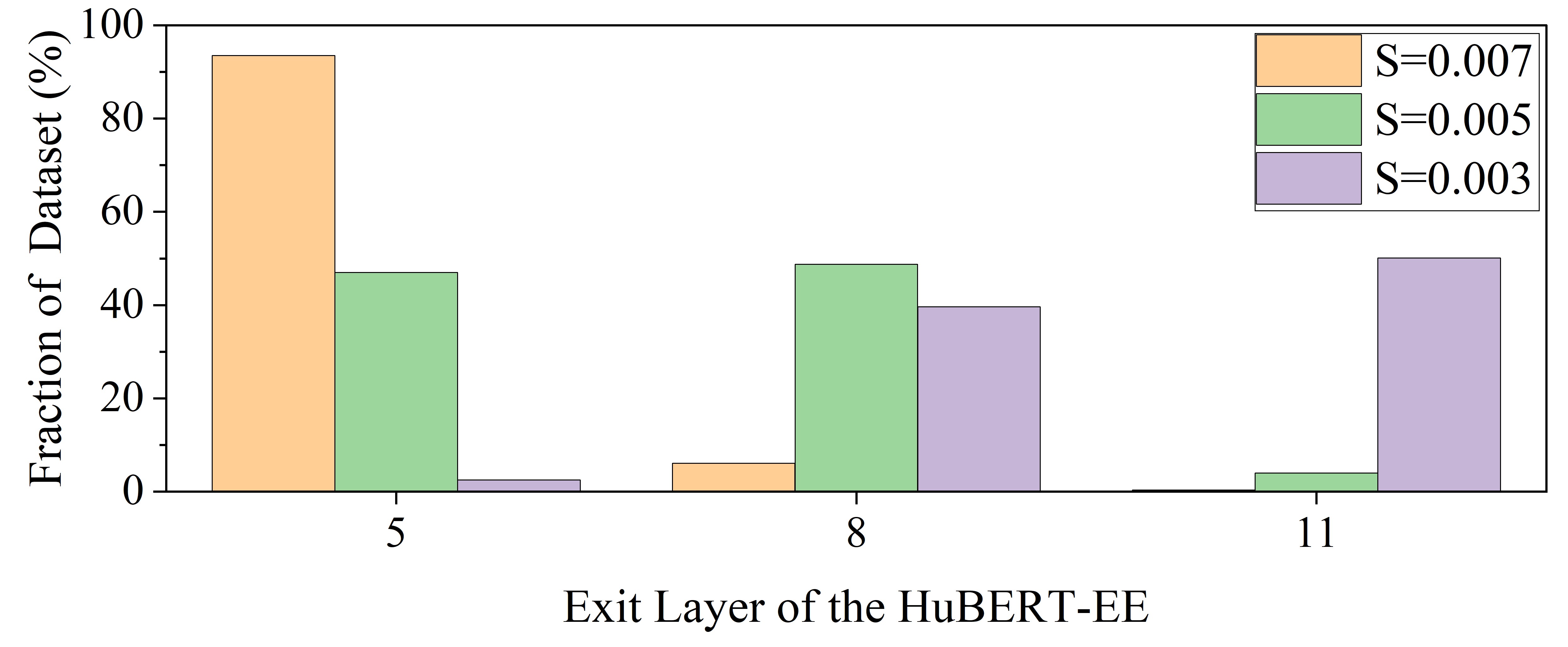}
     \vspace{-3mm}
	\caption{
    The number of exiting samples on text-clean. Samples that did not exit earlier were returned via the last linear layer.}
    
	\label{layerwise}
\end{figure}

We experimentally attached three early exit branches to the intermediate layer: $5^{th}$, $8^{th}$, and $11^{th}$ layers of the HuBERT-base model.
As displayed in Figure \ref{layerwise}, we further showed the distribution of exit layers while varying the entropy threshold $S$ from  \{0.007, 0.005, 0.003\}.
For instance, when the entropy threshold was set to $S=0.007$, approximately 94 \% of the samples completed the inference at the first early exit branch. This indicates that a significant majority of the samples were able to exit early based on the given criterion. The results further demonstrated that as the entropy threshold increased, a larger proportion of samples exited earlier, highlighting the effectiveness of the utterance-level entropy criterion in making early exit decisions for the ASR task.

\section{Limitations}
In our study, we employed an entropy-based metric as the criterion for early exiting. However, we observed that the entropy values were relatively small due to the peak feature of the CTC softmax outputs. As a result, the entropy-based metric became sensitive and required careful selection of an appropriate threshold. 
This was crucial to prevent premature exit or unnecessary computations during the inference process.
Therefore, it is important to consider the specific task and dataset characteristics and carefully choose an appropriate threshold to ensure optimal performance and avoid any potential drawbacks related to early exiting decisions.

\section{Conclusions}
In this paper, we introduced a novel early exit mechanism for ASR, namely HuBERT-EE, that can dynamically accelerate the inference of a large-scale HuBERT model.
From the experimental results on the LibriSpeech, it is verified that the HuBERT-EE was successfully applied to the ASR task while achieving a better quality–efficiency trade-off compared to other compression techniques.
Moreover, we conducted detailed analyses to determine the optimal training strategy and early exit criterion for the early exit branch.

\newpage

\bibliographystyle{IEEEtran}
\bibliography{mybib}

\end{document}